\documentclass{article}
\pdfoutput=1
\usepackage{spconf,amsmath,graphicx}
\usepackage{array,multirow}
\usepackage{enumitem}
\usepackage{cite}
\usepackage{lpic}
\usepackage{microtype}
\usepackage{booktabs}
\usepackage{comment}
\usepackage{hyperref}

\newcommand\blfootnote[1]{%
  \begingroup
  \renewcommand\thefootnote{}\footnote{#1}%
  \addtocounter{footnote}{-1}%
  \endgroup
}
\title{IMPROVING END-TO-END MODELS FOR SET PREDICTION IN \\ SPOKEN LANGUAGE UNDERSTANDING}
\name{\begin{tabular}{c} Hong-Kwang J. Kuo, Zolt{\'a}n T\"{u}ske$^{*}$, Samuel Thomas, Brian Kingsbury, George Saon \end{tabular}}
\address{\begin{tabular}{c} IBM Research AI \end{tabular}}

\begin{document}
\ninept
\newcommand{\argmax}{\operatornamewithlimits{argmax}}
\maketitle
\begin{abstract}
  The goal of spoken language understanding (SLU) systems is to determine the meaning of the input speech signal, unlike speech recognition which aims to produce verbatim transcripts. Advances in end-to-end (E2E) speech modeling have made it possible to train solely on semantic entities, which are far cheaper to collect than verbatim transcripts.  We focus on this set prediction problem, where entity order is unspecified.  Using two classes of E2E models, RNN transducers and attention based encoder-decoders, we show that these models work best when the training entity sequence is arranged in spoken order. To improve E2E SLU models when entity spoken order is unknown, we propose a novel data augmentation technique along with an implicit attention based alignment method to infer the spoken order.  F1 scores significantly increased by more than 11\% for RNN-T and about 2\% for attention based encoder-decoder SLU models, outperforming previously reported results.
\end{abstract}
\blfootnote{*Work done while at IBM. Currently at AppTek.}

\section{Introduction}
\label{sec:intro}

SLU systems have traditionally been a cascade of an automatic speech recognition (ASR) system converting speech into text followed by a natural language understanding (NLU) system that interprets the meaning of the text~\cite{mesnil2014using,raymond2007generative,liu2016joint,huangadapting}. In contrast, an end-to-end (E2E) SLU system~\cite{Haghani2018,ghannay2018end,lugosch2019speech,caubriere2019curriculum,huang2020leveraging,palogiannidi2020end,radfar2020end,rongali2020exploring,agrawal2020tie,morais2020end} processes speech input directly into meaning without going through an intermediate text transcript.

\begin{figure}
    \centering
    \includegraphics[width=0.85\columnwidth]{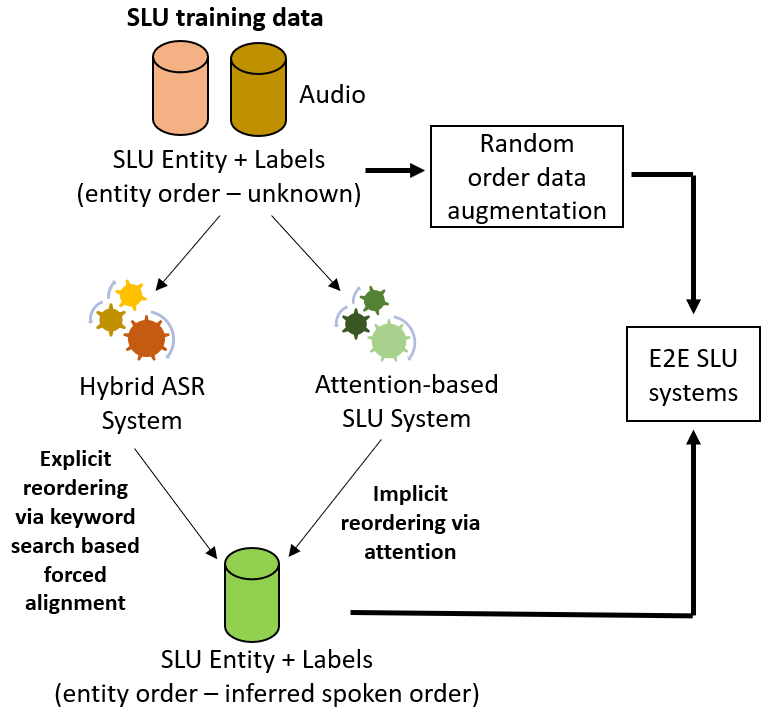}
    \caption{Training E2E SLU models when entity spoken order is unknown via data augmentation and entity reordering.}
    \label{fig:arch}
\end{figure}

End-to-end sequence-to-sequence models can flexibly be trained on different types of ground truth. For ASR, the training data is speech with verbatim transcripts, shown as example (0) below.  To train an SLU model, sentences need to be annotated with entity labels plus a label representing the intent of the entire utterance, as shown in (1).  In (2), entities are presented in natural spoken order, but words not belonging to entities are excluded.  SLU can be considered a set prediction task, as the meaning depends only on the set of semantic entities, not the order in which they were spoken. (3) represents a condition where the spoken order of the entities is unknown, so their order is standardized via lexicographic sorting on label names (e.g. stoploc.city\_name).  
\begin{enumerate}[leftmargin=*]
\setcounter{enumi}{-1}
    \item {\bf Transcript:} {\normalsize i want a flight to dallas from reno that makes a stop in las vegas }

    \item {\bf Transcript+entity labels:}  {\normalsize {\it i want a flight to} DALLAS \mbox{\normalsize B-toloc.city\_name} {\it from} RENO \mbox{\normalsize B-fromloc.city\_name} {\it that makes a stop in} LAS \mbox{\normalsize B-stoploc.city\_name} VEGAS \mbox{\normalsize I-stoploc.city\_name} \mbox{\normalsize INTENT-flight} }
    
    \item {\bf Entities in natural spoken order:} {\normalsize DALLAS \mbox{\normalsize B-toloc.city\_name} RENO \mbox{\normalsize B-fromloc.city\_name} LAS \mbox{\normalsize B-stoploc.city\_name} VEGAS \mbox{\normalsize I-stoploc.city\_name} \mbox{\normalsize INTENT-flight} }
    
    \item {\bf Entities in alphabetic order:}  {\normalsize RENO \mbox{\normalsize B-fromloc.city\_name} LAS \mbox{\normalsize B-stoploc.city\_name} VEGAS \mbox{\normalsize I-stoploc.city\_name} DALLAS \mbox{\normalsize B-toloc.city\_name} \mbox{\normalsize INTENT-flight} }
\end{enumerate}
In this paper we focus on training SLU models on ground truth that is a set of semantic entities with unknown spoken order. This type of data cannot be used to train classical ASR or NLU models, yet it may be abundant and much less costly to collect.  Imagine recording a human agent talking with a client to make a travel reservation, along with the actions performed by the agent, e.g. filling out web forms or other database transaction records which can be translated into semantic entities.  Alternatively, such data may be collected automatically by spoken conversational systems over multiple turns of dialogue through implicit or explicit confirmation of entities.  To train ASR and NLU separately, accurate verbatim transcription of speech data requires 5-10$\times$ real-time for a human transcriber, plus additional costs for labeling entities.  In contrast, automatic collection of speech with an associated set of important entities may be performed in the course of helping the customer and incurs no additional cost.  Training on such data is an important research problem, but is also difficult.

 An important class of E2E models being deployed in many commercial speech services with streaming capabilities is RNN Transducer (RNN-T) models.  Prior work shows RNN-T based SLU can be trained with ground truth containing entities in spoken order, but when the spoken order is unknown, the performance suffers more than 10\% absolute drop in F1 score~\cite{sam2021rnn}.  Attention based encoder-decoder models do better, but F1 still decreases by more than 2\% when spoken order is unknown~\cite{kuo2020end}.

In this paper, as shown in Figure~\ref{fig:arch}, we propose two methods to address this problem: (1) a novel data augmentation scheme using randomly ordered semantic entities for SLU, and (2) a novel method to align semantic entities to speech and infer the spoken order by using the attention values from an attention-based encoder-decoder SLU model.  
With our proposed techniques, we show how RNN-T based SLU models can be effectively trained.  F1 scores are significantly increased by more than 11\% for RNN-T and about 2\% for attention based encoder-decoder SLU models, outperforming previously published state-of-the-art results.

\section{E2E Models for SLU}
End-to-end models directly map a sequence of acoustic features to a sequence of symbols without conditional independence assumptions.
We adapt these popular ASR models for SLU.  

\subsection{RNN Transducer model}
 \label{sec:rnnt}
 RNN-T introduces a special BLANK symbol and lattice structure to align input and output sequences.
 The models typically consist of three different sub-networks: a transcription network, a prediction network, and a joint network~\cite{graves2012sequence}. The transcription network produces acoustic embeddings, while the prediction network resembles a language model in that it is conditioned on previous non-BLANK symbols produced by the model. The joint network combines the two embedding outputs to produce a posterior distribution over the output symbols including BLANK. %
  An RNN-T based SLU model is created in two steps: by constructing an ASR model~\cite{he2019streaming,rao2017exploring,li2019improving,shafey2019joint,ghodsi2020rnn} and then adapting it to an SLU model through transfer learning~\cite{sam2021rnn}.
  In the first step, the model is  pre-trained on large amounts of general purpose ASR data to allow the model to effectively learn how to transcribe speech into text. Given that the targets in the pre-training step are only graphemic/phonetic tokens, prior to the model being adapted using SLU data, it is extended to model the semantic labels.  These new SLU labels are integrated by resizing the output layer and the embedding layer of the prediction network to include additional symbols. The new network parameters are randomly initialized, while the remaining parts are initialized from the pre-trained network. Once the network has been modified, it is subsequently trained on SLU data in steps similar to training an ASR model~\cite{george2021rnn}.
 
 \subsection{Attention based LSTM encoder-decoder model}
\label{sec:attn}
This model estimates sequence posterior probabilities without introducing any explicit hidden variables.
The alignment problem is handled internally by squashing the input stream dynamically with a trainable attention mechanism synchronized to the output sequence.
The model is able to handle problems with non-monotonic alignment exceptionally  well, and has become the state-of-the-art approach in many machine learning problems.
The structures of an RNN-T and attention encoder-decoder model are similar.
The attention based model also contains an LSTM based encoder network to generate acoustic embeddings.
 The single-head LSTM decoder contains a language model like component, and the attention module which combines the acoustic embeddings and the embeddings of the symbol sequence into a context vector to predict the next symbol.
The adaptation of attention based encoder-decoder ASR models\cite{tuske2020} to SLU~\cite{kuo2020end} can be carried out using the same steps as described for RNN-T.

 \subsection{Attention based Conformer encoder-decoder model}
\label{sec:conformer-attn}

For the model described in the previous section, only the decoder contains an attention mechanism.  We also consider adding attention to the encoder.  A conformer is a combination of convolutional neural network and self-attention based transformer which has been shown to achieve state-of-the-art speech recognition results~\cite{gulati2020conformer}.   We investigate an attention model where the encoder is a conformer, which in a recent study~\cite{tuske2021} achieved a slight improvement over an LSTM based encoder for the Switchboard 300h ASR task. The study found no benefit in replacing the decoder with a conformer, so we do not change the single-head LSTM decoder for our attention model.

 \section{Set Based Data Augmentation}
 \label{sec:augment}

For the set prediction problem, we are provided with a set of entities without knowing the spoken order. To train our sequence-to-sequence model, we had arbitrarily chosen to standardize the entity order by alphabetic sorting.  To improve robustness, we propose data augmentation that randomizes the order of the entities in the ground truth that is used to pre-train various E2E models.
During this pre-training phase, the model is presented with a different version of ground truth at each epoch. This is followed by a fine-tuning phase where the model is trained on ground truth with entities in alphabetic order. Exposing the model in the pre-training phase to many examples with entity order mismatch between ground truth and speech may help it to learn better during fine-tuning.

\section{Reordering Sets into Spoken Order}
\label{sec:align}

\subsection{Explicit keyword search based alignment}
\label{sec:alignHybrid}
In our first method to find the underlying spoken order of the set of entities, we employ a simple procedure similar to keyword spotting. %
In acoustic keyword spotting, a combination of two kinds of acoustic models is used.  While the keyword being searched for is modeled by its underlying phonetic string, all non-keyword speech is modeled by a garbage model. %
Using a conventional hybrid ASR model, we construct a model for the keyword being searched as a concatenation of the hidden Markov models (HMMs) corresponding to the constituent phones in the keyword. The garbage model is represented by a generic phone for vocal speech. We then concatenate these models: first the garbage model, then the keyword model, and finally the garbage model again, and then force-align the utterance to the keyword model using the hybrid ASR model.

\subsection{Implicit internal alignment using attention} %
\label{sec:alignAttn}

\begin{figure}[t]
\includegraphics[width=\columnwidth]{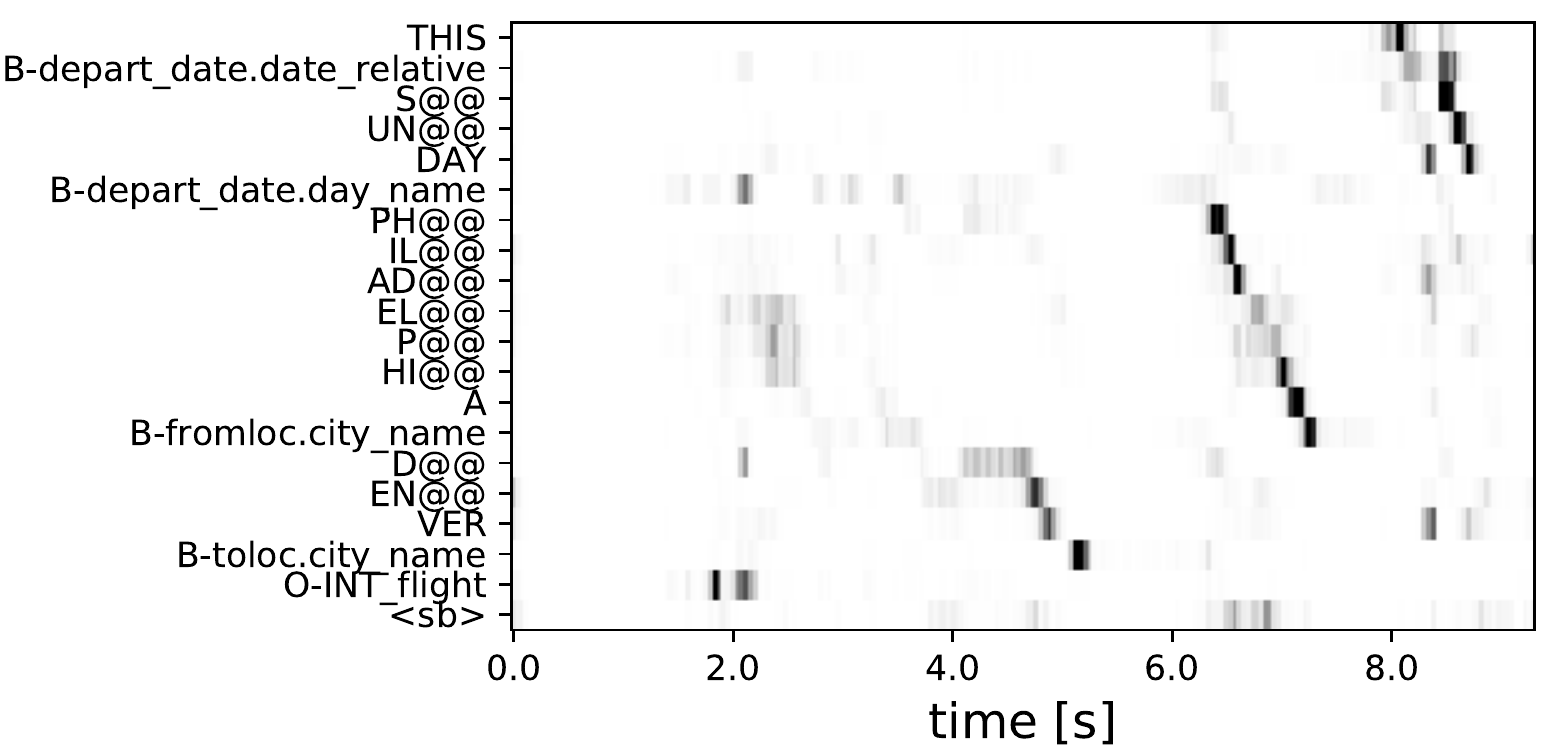}
\caption{Attention plot for ``I would like  to  make  a reservation for a flight to Denver from Philadelphia on this coming Sunday'' where ground truth is entities in alphabetic order by label name
}
\label{fig:attn}
\vspace{-3mm}
\end{figure}

As has been observed in \cite{kuo2020end}, an attention model can handle SLU entities in non-spoken order, and a single-head attention can have an especially sharp focus for spoken tokens at the corresponding time-position in the acoustic feature stream.  
Based on this observation, the spoken order of SLU phrases can be estimated.
The following heuristic estimates an average time position for each SLU phrase when the spoken order of the phrases is unknown:
\begin{align}
\label{eq:attalign}
  t_i = \frac{1}{|\mathcal{N}_i|} \sum_{n \in \mathcal{N}_i} \argmax_t \alpha_{t,n}
\end{align}
where $\alpha_{t,n}$ denotes the attention for the $n$th output token at each acoustic frame $t$.
Let the $i$th SLU phrase, consisting of spoken BPE (byte pair encoding) tokens~\cite{subword-nmt} and entity labels, start at position $n_i$ and end at \mbox{$n_{i+1}-1$} in the output sequence, and let $\mathcal{N}_i$ contain only the positions of the BPE (spoken) tokens.
Figure~\ref{fig:attn} shows an example attention plot where the x-axis is time within the speech signal (corresponding to $t$) and the y-axis contains the sequence of BPE tokens and entity labels (corresponding to $n$, from top to bottom), and the value of $\alpha_{t,n}$ is represented by how dark the pixel is.
Considering only spoken tokens, Eq.~\ref{eq:attalign} thus calculates an average time-position for each SLU phrase, by which the spoken order of the phrases can be reestablished.

This alignment method is better than using hybrid ASR because the SLU model can first be adapted acoustically by training on SLU data where the ground truth is a set of semantic entities with unknown spoken order.  Adapting the hybrid ASR on such data is not as straightforward.

\section{Experimental Setup}
\subsection{Dataset}
\label{sec:dataset}
ATIS (Air Travel Information Systems)~\cite{hemphill1990atis} is a publicly available Linguistic Data Consortium (LDC) corpus that has been widely used in SLU research. We use the same data preparation as~\cite{kuo2020end,sam2021rnn}.  There are 4976 training audio files ($\sim$9.64 hours, 355 speakers) and 893 test audio files ($\sim$1.43 hours, 355 speakers) downsampled to 8~kHz. To better train E2E models, additional copies of the corpus are created using speed/tempo perturbation, resulting in $\sim$140 hours for training. To simulate real-world operating conditions, we create a second \textit{noisy} ATIS corpus by adding street noise between 5-15~dB signal-to-noise ratio (SNR) to the clean recordings. This $\sim$9.64 hour noisy data set is also extended via data augmentation to $\sim$140 hours. A corresponding noisy test set is also prepared by corrupting the original clean test set with additive street noise at 5~dB SNR.

Intent recognition performance on this dataset is measured by intent accuracy, while semantic entity recognition (slot filling) performance is measured with the F1 score~\cite{kuo2020end}. 

\subsection{RNN-T SLU model}
As described in \cite{sam2021rnn}, the RNN-T models we develop for SLU are first pre-trained on task independent ASR data.  In this paper we use an  ASR  model trained  on 300 hours of data from the Switchboard corpus using the steps described in~\cite{george2021rnn}.
CTC acoustic models are first trained and used to initialize the  transcription network of the RNN-T model~\cite{audhkhasi2019forget,kurata2019guiding}.
The RNN-T model used in our experiments has a transcription network containing 6 bidirectional LSTM layers with 640 cells per layer per direction. The prediction network is a single unidirectional LSTM layer with 768 cells. 
The  joint  network  projects  the  1280-dimensional stacked encoder vectors from the last layer of the transcription net and the 768-dimensional prediction net embedding  each  to  256  dimensions,  combines  them  multiplicatively,  and  applies  a  hyperbolic  tangent.
After this, the output is projected to 46 logits, corresponding to 45 characters plus BLANK, followed by a softmax layer. In total, the model has 57M parameters.
The models were trained in PyTorch for 20 epochs \cite{pytorch}.
More details of the design choices can be found in \cite{george2021rnn}. As described earlier, during SLU adaptation, new network parameters are randomly initialized while the remaining parts of the network are copied from the pre-trained  network.  For the ATIS entity/intent task we add 151 extra symbols to the joint network output layer and the prediction network input embeddings as entity/intent label targets.

\subsection{Attention based LSTM encoder-decoder SLU model}
Our attention based E2E model follows the structure of \cite{tuske2020}.
It has a 6-layer bidirectional LSTM encoder and 2-layer unidirectional LSTM decoder, and models the posterior probability of about 600 BPE units augmented with the entity and intent labels.
All LSTM layers use 768 nodes per direction.
The first LSTM of the decoder operates only on the embedded predicted symbol sequence, while the second LSTM processes acoustic and symbol information using a single-head additive location-aware attention mechanism \cite{chorowski15}.
The dropout and drop-connect rates are set to 0.3 in the encoder and to 0.15 in the decoder.
In addition zoneout with 0.10 probability is also applied in the second LSTM layer of decoder \cite{Krueger2017}.
Overall, the model contains 57M parameters.
For ASR pretraining the standard Switchboard-300 corpus is used, and the model is optimized from random initialization by AdamW in 450k update steps with a batch of 192 sequences \cite{loshchilov2018decoupled}.
The SLU fine-tuning is carried out with a batch of 16 sequences in about 100k steps.

\subsection{Attention based Conformer encoder-decoder SLU model}
To add self attention to the encoder, we replace the LSTM encoder with a Conformer encoder, following the structure of~\cite{tuske2021}. Everything else is the same.  Overall, the model contains 68M parameters.

\section{Results}

Table~\ref{tab:ATISrnnt} shows results with RNN-T models on the ATIS clean corpus described in Section~\ref{sec:dataset}. Using the unadapted baseline ASR, the WER is 14.0\%. When full verbatim transcripts with semantic labels are used to adapt the ASR model into an SLU model, we obtain the results in [R1c].  
[R2c] shows results when ground truth containing only entities in natural spoken order are used for SLU model training.  Compared with using full transcripts [R1c], we see hardly any difference in F1 (93.3\% vs. 93.0\%), despite the fact that [R2c] has an astonishing WER of 60.6\%.  The WER is so high because the model does not output non-entity words that do not directly impact the meaning representation. These results show that accurate SLU models can be trained without full transcripts.    
During training of [R3c] the ground truth entities are given without information about the spoken order and are sorted alphabetically based on the name of the entity label. Compared with [R2c], F1 suffers a huge degradation of 11.8\% (from 93.0\% to 81.2\%). Although RNN-T models are currently one of the most popular speech recognition models, such a transducer model does poorly when the ground truth sequence does not correspond to the spoken sequence because it operates monotonically. The approach proposed in this paper addresses this issue.  

\begin{table}[t]
  \caption{{\rm ATIS} WER and set-of-entities slot filling F1 score for speech input using RNN-T models}
  \label{tab:ATISrnnt}
  \centering
  \begin{tabular}{lcc@{}} \toprule
    {\bf Training Data} &  {\bf WER} & \bf{F1}  \\  \cmidrule(lr){1-1}
    \cmidrule(lr){2-3}
    \textbf{Clean speech} \\
     $[$R0c$]$ Base ASR                     & 14.0  &      \\ 
    $[$R1c$]$ Full transcripts              &  1.4  & {\bf 93.3} \\ 
    $[$R2c$]$ Entities, spoken order        & 60.6  & 93.0 \\
    $[$R3c$]$ Entities, alphabetic order    & 73.1  & {\bf 81.2} \\
    \bottomrule
  \end{tabular}%
\end{table}

\begin{table}[t]
  \caption{{\rm ATIS} set-of-entities slot filling F1 score for speech input using RNN-T [R], attention based encoder-decoder with LSTM [L] encoder or Conformer [C] encoder}
  \label{tab:ATISfull1}
  \centering
  \centering
  \begin{tabular}{lccc@{}}
    \toprule
    {\bf Training Data}                    & {\bf [R]} & \bf{[L]} & \bf{[C]} \\
    \cmidrule(lr){1-1}                       \cmidrule(lr){2-4}
    \textbf{Clean speech} \\
    $[$1c$]$ Full transcripts              & 93.3 & 94.4 & {\bf 94.3} \\ 
    $[$2c$]$ Entities, spoken order        & 93.0 & 94.2 & 94.3 \\
    \hline
    $[$3c$]$ Entities, alphabetic order    & {\bf 81.2} & 92.0 & 92.7 \\
    $[$4c$]$ Random order augmentation     & 79.4 & 92.7 & 93.1 \\
    $[$5c$]$ Spoken order alignment-H      & 92.9 & 93.6 & 93.7 \\
    $[$6c$]$ Spoken order alignment-A      & 92.8 & 93.4 & 93.8 \\
    $[$7c$]$ + Random order augmentation   & {\bf 92.6} & 94.0 & {\bf 94.3} \\
    \hline \hline
    \textbf{Noisy speech} \\
    $[$1n$]$ Full transcripts              & 92.0 & 93.2 & 93.4 \\ 
    $[$2n$]$ Entities, spoken order        & 90.8 & 92.2 & 93.2 \\
    \hline
    $[$3n$]$ Entities, alphabetic order    & 70.7 & 89.5 & 90.9 \\
    $[$4n$]$ Random order augmentation     & 81.7 & 90.3 & 91.6 \\
    $[$5n$]$ Spoken order alignment-H      & 89.1 & 90.8 & {\bf 91.5} \\
    $[$6n$]$ Spoken order alignment-A      & 89.2 & 91.4 & {\bf 92.9} \\
    $[$7n$]$ + Random order augmentation   & 89.5 & 92.0 & 93.0 \\
    \bottomrule
  \end{tabular}%
\end{table}

Focusing on just entity recognition, Table~\ref{tab:ATISfull1} shows F1 score results comparing the three E2E SLU models: RNN-T [R], attention based encoder-decoder with LSTM [L] encoder or Conformer [C] encoder.  The F1 score is around 93-94\%.  Although not shown in the table, the intent accuracy is around 96-97\% for the attention models.  These results are comparable with or better than previously reported results~\cite{cao2020style,huangadapting,liu2016joint,mesnil2014using} on ATIS for speech input. In our experiments, the intent accuracy did not vary much with different types of training transcripts (e.g. [1c], [2c], [3c]), so we do not report this metric in the rest of the paper.  In addition to the clean corpus, results are also shown using the noisy corpus as described in Section~\ref{sec:dataset}. 

With clean speech, F1 scores when trained on just entities in spoken order [2c] are similar to those with full transcripts [1c].   When the entity spoken order is unknown [3c], while RNN-T suffers a big loss (12\%) in F1, the attention based encoder decoder models do relatively well, but F1 still decreases by about 2\%. Similar trends are observed for the noisy test set.

\subsection{Improving Set Prediction}

Lines [4c-7c]/[4n-7n] in Table~\ref{tab:ATISfull1} show results of experiments to compensate for F1 degradation when the entity spoken order is unknown.

First we apply data augmentation as described in Section~\ref{sec:augment}, where we expose the model in a pre-training phase to ground truth with entities in various random orderings, followed by fine-tuning on alphabetic order entities. Results are shown in lines [4c]/[4n].  Small but consistent improvements are observed for the attention based encoder-decoder models, while the results for RNN-T are inconsistent. To verify that the gains are due to data augmentation and not more epochs of training or resetting the learning rate, we did a control experiment where alphabetic order entities were used in both pre-training and fine-tuning phases, and did not see any improvement. 

As described in Section~\ref{sec:align}, next we infer the spoken order of the entities by aligning the entities to the speech, and then use this ground truth to train the SLU model.  We used two methods for alignment, one based on a hybrid ASR model (Section~\ref{sec:alignHybrid}), with results shown in [5c]/[5n], and the other based on an attention model (Section~\ref{sec:alignAttn}, using the models in [L3c]/[L3n] to compute attention values), with results shown in [6c]/[6n].  For both alignment methods, the entity alignment error of clean training data is relatively small, 3\% for hybrid and 2\% for attention. For noisy data, the alignment error increases to 6\% for hybrid but remains 2\% for attention.  The attention model used to infer spoken order is [L3n], which has been adapted to noisy in-domain speech data based on ground truth of entities in alphabetic order.  This is an advantage of implicit alignment using an attention SLU model, since it is not straightforward to adapt the hybrid model on the noisy speech data without verbatim transcripts.  The better alignment quality translates into better final SLU models: [6n] results are consistently better than [5n], e.g. for the Conformer encoder model, [C6n](F1=92.9\%) is better by 1.4\%.  

Finally, in [7c]/[7n], we apply both methods, where we initialize with a model pre-trained on randomly ordered entities and apply fine-tuning on re-ordered ground truth as in [6c]/[6n].  Compared to the baseline [3c], RNN-T performance increased by 11.4\% from 81.2\% to 92.6\% and attention models increased by about 2\%.  In fact, for the Conformer encoder model, [C7c](F1=94.3\%) is identical to that trained on full transcripts [C1c]. In the noisy condition, RNN-T improved from 70.7\% to 89.5\% (18.8\% increase), and data augmentation consistently helps, even accounting for the reorganization of entities into spoken order.

\section{Conclusions}

End-to-end models for speech recognition are flexible models that can be adapted into SLU models that directly decode speech into a meaning representation such as a set of semantic entities and sentence intent. In this paper, we addressed how best to train such an SLU model to perform this set prediction, given training data where the spoken order of entities is unknown. We investigated two methods to improve performance: data augmentation with randomly ordered entities and pre-aligning the entities with speech to put them in spoken order for training. The data augmentation is novel in that it is applied at the output label level instead of input feature level.  The alignment method is novel because it relies on an attention based SLU model, which has distinct advantages over a hybrid ASR model since it can first be adapted to noisy speech data with non-verbatim ground truth, resulting in 3 times lower alignment error (2\% vs. 6\%). Together, our proposed methods improved SLU performance by 11.4\% for RNN-T and about 2\% for attention models. For the best model, the attention model with Conformer encoder, performance is improved to a level similar to being trained on full transcripts.  In summary, we have demonstrated the ability to train on SLU data associated with a set of semantic entities with unknown spoken order.  Such data can automatically be collected without needing human supervision, significantly reducing the cost of building E2E SLU systems, while achieving performance similar to that with full transcripts.

\vfill\pagebreak
\bibliographystyle{IEEEbib-abbrev}
\bibliography{refs}

\end{document}